\documentclass[12pt]{article}

\usepackage[pdftex]{graphicx}
\usepackage{amssymb}
\usepackage{amsmath}
\usepackage{tabularx}
\usepackage{fourier}
\usepackage{hyperref}
\hypersetup{
  colorlinks,
  citecolor=Green,
  linkcolor=Red,
  urlcolor=Blue}
\usepackage{multicol}
\usepackage{listings}
\usepackage{authblk}
\usepackage{setspace}
\usepackage[table,usenames,dvipsnames]{xcolor}
\usepackage{upgreek}
\usepackage{longtable}
\usepackage{varwidth}
\usepackage{multirow}
\usepackage{pbox}
\usepackage[boxed]{algorithm2e}
\usepackage{spverbatim}

\newcommand{\indeg}[0]{\mathrm{k}}
\newcommand{\outdeg}[0]{\mathrm{k'}}
\newcommand{\du}[0]{\mathrm{d}}
\newcommand{\dd}[0]{\mathrm{d_D}}
\newcommand{\dr}[0]{\mathrm{d_R}}
\newcommand{\aff}[0]{\uppsi}

\renewcommand\emph[1]{{\textit{#1}}}

\title{Symbolic regression of generative network models}

\author
{Telmo Menezes,$^{*1,2}$ Camille Roth$^{1}$\\
\normalsize{CNRS\\$^{1}$Centre Marc Bloch Berlin (An-Institut der Humboldt Universit\"at, UMIFRE CNRS-MAE)}\\
\normalsize{Friedrichstr. 191, 10117 Berlin, Germany}\\
\normalsize{$^{2}$Centre d'Analyse et de Math\'ematique Sociales (UMR 8557 CNRS-EHESS)}\\
\normalsize{190 av. de France, 75013 Paris, France}\\
}

\date{}

\begin{document}

\maketitle

\begin{abstract}

Networks are a powerful abstraction with applicability to a variety of scientific fields. Models explaining their morphology and growth processes permit a wide range of phenomena to be more systematically analysed and understood. At the same time, creating such models is often challenging and requires insights that may be counter-intuitive. Yet there currently exists no general method to arrive at better models. We have developed an approach to automatically detect realistic decentralised network growth models from empirical data, employing a machine learning technique inspired by natural selection and defining a unified formalism to describe such models as computer programs. As the proposed method is completely general and does not assume any pre-existing models, it can be applied ``out of the box'' to any given network. To validate our approach empirically, we systematically rediscover pre-defined growth laws underlying several canonical network generation models and credible laws for diverse real-world networks. We were able to find programs that are simple enough to lead to an actual understanding of the mechanisms proposed, namely for a simple brain and a social network.
\end{abstract}

\section{Introduction}

Increasingly many scientific domains rely on the concept of networks to represent an observable state of a system, where networks are usually seen as the outcome of a generative process.
For systems without centralised control, these generative processes consist of local interactions between entities, be they proteins, neurons, organisms, people or organisations.

While current technological advances have been making it increasingly easy to collect datasets for large networks, it is difficult to extract 
models from this data. This difficulty can be attributed both to the sheer size of the datasets and to the non-linear dynamics of many of these decentralised systems, which resist reductionist methodologies. Another difficulty is posed by the mapping between generative models and observable networks since there is a many-to-many correspondence between generative models and observable networks. A network may be explained by different models and a model -- provided it is stochastic in nature -- may be capable of generating different classes of networks due to the amplification of initial random fluctuations.

Following conventional scientific methodology, researchers devise
models that can account for a network and then test the quality of the
model against a number of metrics. Much-cited examples include
preferential attachment \cite{bib1}, competition between nodes \cite{bib2, bib3}, team assembly mechanisms \cite{bib4}, random networks with constraints\cite{bib5, bib6, bib7}, \emph{inter alia}. Models are typically based on intuition or prior evidence that such and such process appears to be particularly important in the formation of interactions.  A problem here is that of human bias in looking for good models. There is always the possibility that high-quality models are counter-intuitive, and thus unlikely to be proposed by researchers.

The work we report in this paper work is aligned with the idea of
creating \emph{artificial scientists}. Parts of the scientific method
are automated, namely the generation and refinement of hypothesis, as
well as their testing against observables. For example, in a work with
some parallels to the ideas presented in this paper, scientific laws
are extracted from experimental data using genetic programming \cite{bib8}.

There have been some preliminary attempts at using genetic programming
to search for network models \cite{bib21, bib22, bib23}, and to structural analysis and community detection \cite{bib24}. However, to the best of our knowledge, we provide the first proof-of-concept application of symbolic regression to discover and select plausible morphogenetic processes for real-world networks. The method we propose can be applied to both synthetic networks and on real-world networks. In the case of synthetic networks, it makes it possible to discover the exact generative rule used to construct the particular type of network in question, while in the case of real-world networks, it proposes a generative rule that robustly reproduces the original topological features. Furthermore, in contrast with previous works, our approach relies only on local information and uses a parameter-free fitness function without any \emph{ad-hoc} assumptions. It eventually provides a straightforward mapping to mathematical expressions.  A more detailed comparison to \cite{bib22, bib23} is provided in the \hyperlink{sup_mat}{supplemental materials}.

\section{Results} 

\subsection{Generator search}
Machine learning techniques can be used to help researchers generate
alternative models that are capable of reproducing networks with
certain topological features. The approach we propose employes
\emph{genetic programming} \cite{bib13, bib14}, a form of \emph{evolutionary computation}. Genetic programming is a type of search inspired by natural selection where evolutionary pressure is created to guide a population of solutions to increasingly higher quality. In this case the individuals in the population are network generative models, and the quality measure is how much a synthetic network generated by a model approximates the real observable network.

Two fundamental issues have to be addressed in implementing this technique. Firstly, the models need to have a representation that is uniform and permits recombination. Secondly, an appropriate measure of similarity needs to be defined so that synthetic and real-world networks can be compared.

The first issue touches on a shortcoming in the current literature on ``network science'': there is no unified and elegant way of formally representing network generative processes.  To address this we introduce the concept of \emph{network generator as a computer program} which, for the purposes of this article, we refer to simply as \emph{generators}. We define a network generative process as a sequence of discrete steps where a new arc is created at each step. The process can be straightforwardly applied to both directed and undirected networks. At any given moment, there is a set of possible arcs that could be created. A generator becomes fully defined if it provides a way to prefer some {arc} over the others. Instead of attempting to define a deterministic selection process we create a stochastic one --- recognising that many of the generative processes that produce networks have some degree of intrinsic randomness.

The generator is thus a function $w(i,j)$ that assigns a weight $w_{ij}$ to all arcs $(i,j)$ from a random sample $S$ (see \hyperlink{methods}{Methods}). At each network construction step, a new arc is then stochastically selected with a probability $P_{ij}$ proportional to $w_{ij}$ such that:
\begin{equation}
P_{ij} = \frac{w'_{ij}}{\displaystyle\sum_{(i',j') \in S} w'_{i'j'}}
\end{equation}

where $w'_{ij} = w_{ij}$ if $w_{ij} > 0$, $0$ otherwise. If all the weights for a sample are zero, they are all set to $1$ to avoid division by zero in the above probability expression.

\begin{figure}
\begin{center}
\includegraphics[scale=0.6]{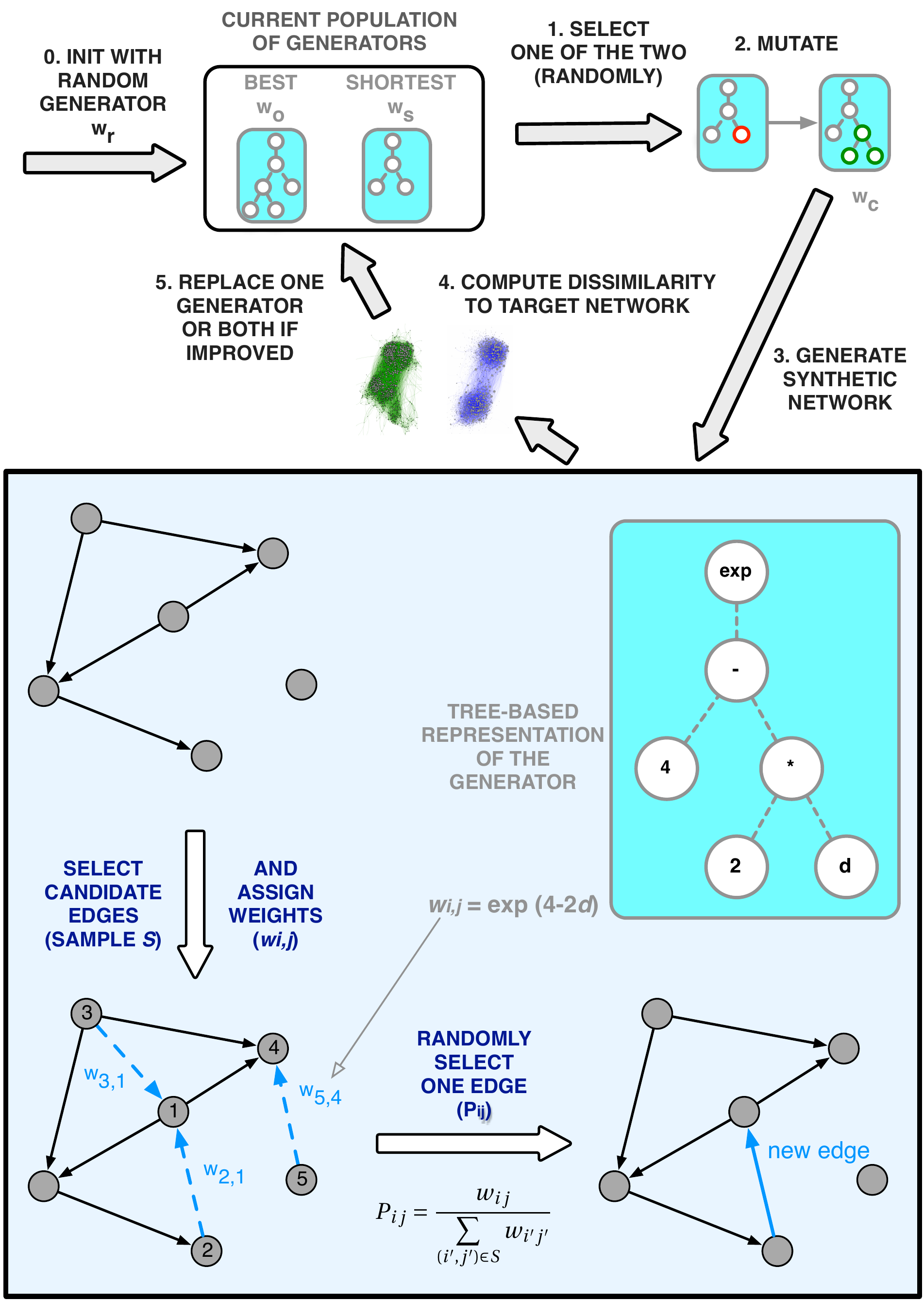} 
\end{center}
\caption{{\bf Automatic discovery of models.} Evolutionary loop including the synthetic network generation process. The top part of this figure describes evolution at the generator population level, while the bottom (framed) part describes the evolution of a network for a given generator.}
\label{fig1}
\end{figure}

The core of our approach then consists in designing a process able to automatically discover weight computation functions $w$ which produce realistic networks.
Generators are represented as \emph{tree-based computer programs}, which are equivalent to mathematical expressions. Tree leaves are variables and constants, and its other nodes are operators. These are the building blocks of our generators (see figure~\ref{fig1}). The set of available operators includes simple arithmetic operators: $\{+$, $-$, $*$, $/\}$, general-purpose mathematical functions: $\{x^y$, $e^x$, $\log$, $\text{abs}$, $\min$, $\max\}$, conditional expressions: $\{>, <, =, =0\}$ and an affinity function ($\aff$). Variables contain information specific to the two vertices participating in the {arc}: in- and out-degrees ($\indeg$ and $\outdeg$), undirected, directed and reverse distances between the two vertices ($\du$, $\dd$ and $\dr$) and their sequential identifiers ($i$ and $j$). In the case of undirected networks only $\indeg$, $\du$, $i$ and $j$ are used. Sequential identifiers and the related affinity function will be discussed later on.

We rely on a random walk-based heuristic distance: not only would the explicit computation of all exact pairwise distances during the generative process be too computationally expensive, but perhaps more importantly, new connections are also likely to be accurately construed as a hop-by-hop navigation mechanism instead of a selection process based on an omniscient distance value (see Supp. Info.).

This simple arrangement configures a uniform language to describe generators capable of expressing entity-level behaviours that produce non-linear, non-centralised network growth processes.

\bigskip
The second issue of measuring network similarity is addressed by
comparing a set of conventional features of both networks. We combine
distributions that describe {simple} aspects of the network, such as
in- and out- degree, {direct and reverse} PageRank \cite{bib9}
centralities ({considering actual and, respectively,} inverted arcs),
with {distributions describing finer and more meso-level aspects of
  the structure}, such as  directed/undirected distances and triadic
profiles \cite{bib10}.

These features are reduced to metrics by computing dissimilarities
between the respective distributions. We rely on two notions of
distribution dissimilarities. For degree and \hbox{PageRank}
centralities we apply the \emph{Earth mover's distance (EMD)} \cite{bib11}, for the more sophisticated distance distributions and triadic profiles we rely on a simpler ratio-based dissimilarity metrics (see Supp. Material for a longer discussion). These dissimilarity metrics allow us to determine whether we are converging towards the original distributions at a small computational cost  (other dissimilarity metrics may well be used, but we found these to work well in our case).

We are interested in minimising all of the dissimilarity measures to get as close as possible to the target (real) network. This configures a multi-objective optimisation problem with possible trade-offs since some dissimilarities might need to be minimised at the expense of others. Our objective is to find a balanced solution and employ the following simple strategy: we decide to place all metrics on the same scale and configure their meaning as the improvement with respect to a random network. In practice, each dissimilarity between the target network and a candidate network is divided by the mean dissimilarity between the target network and $30$ Erd\H{o}s-R\'enyi (ER) random networks with the same number of vertices and arcs as the target. For a given metric, this means that if the dissimilarity between the target network and the ER average is, say, 5 and the distance from the target network to the candidate network is 3, the ratio is 3/5. A ratio of 1 thus corresponds to no improvement.
The evolutionary algorithm then tries to improve models by minimising the highest of these ratios, which thereby defines a fitness function.

While ER is assuredly a basic null model, opting for a more sophisticated model may induce undesired bias: for instance, using the configuration model would precisely incorporate the degree distributions of the target network, making it impossible to directly approximate it using the fitness function.

A further feature of our framework is to not assume homogeneity between nodes, irrespective of their structural position. A heterogeneous model is one that starts with the assumption that not all entities in the system behave the same. For example, in a social network, some agents might be intrinsically more likely to form ties. Or they might be more likely to interact within a specific class of agents. We introduce heterogeneity by way of the sequential identifier input variable $i \in \{1,..,n\}$.
These indices, considered as \emph{identifiers}, can then be passed by variable to the generator programs, and used to introduce \emph{a priori} distinctions in behaviours. Let us consider a simple example:

\begin{equation}
w(i, j) = \frac{1}{i}
\end{equation}

This equation describes a generator where the probability of an arc is completely determined by the identifier of the origin vertex. It describes a situation where nodes have different \emph{a priori} propensities to originate connections. Furthermore, it tells us that these propensities are distributed {following} a hyperbolic curve. Even though integer identifiers may appear to be a highly simplistic means of introducing heterogeneity, we need to remember that they can be combined with the other building blocks in an infinity of ways. In the below results from real-world networks we can see that some of the generators that were found make use of the indices in various ways.

Indeed, the simplicity of building blocks can be leveraged and used to facilitate the definition of generators where certain vertices have natural affinity for each other. This is the \emph{affinity} function $\psi$, which uses the modulo operation (remainder of the division of one number by another) to divide the sequence identifier space into a number of $g$ groups, returns $a$ if target and origin nodes $i$ and $j$ belong to the same group (i.e. in case of ``affinity''), and $b$ otherwise:

\begin{equation}
    \uppsi(i, j, g, a, b)= 
\begin{cases}
    a,    & \text{if } (i \bmod g) \equiv (j \bmod g)\\
    b,              & \text{otherwise},
\end{cases}
\end{equation}
 
From now on, we will consider $i$ and $j$ to be implicit parameters and write the function simply as: $\uppsi(g, a, b)$.

We now have a methodological framework that we can use to generate
plausible models for network generators. Several runs on the same
target network may generate different models --- although we will show
experimental evidence that they tend to converge on the same
behaviors. It is now up to the researcher to select amongst them,
possibly using his domain knowledge. A more objective consideration is
the trade-off between simplicity and precision. Our representation of
generators allows for a very straight-forward measure of model
complexity: the \emph{program length}. Trivially, the program length
is an upper bound on the \emph{Kolmogorov complexity} \cite{bib15} of the model. This allows us to apply a quantified version of \emph{Occam's Razor}: all other things being equal, choose the model with the lowest program length. In practice, depending on the variations in precision, the researcher might wish to sacrifice some parsimony for some precision, or vice-versa.

\medskip
\subsection{Application to real and synthetic networks}

To assess our method we start by testing if we can discover generators for networks that were produced by generators we defined ourselves. According to our generator semantics, two classical network types can be defined in a very succinct fashion. 

For an ER random network,

\begin{equation}
w_{\text{ER}}(i, j) = c
\end{equation}

where $c$ is any constant value; for a generator based on Preferential Attachment (PA) as in the Barab\'asi-Albert model, 

\begin{equation}
w_{\text{PA}}(i,j) = \indeg(j)
\end{equation}

We used these two generators to produce networks of five different sizes, from $100$ vertices and $1000$ arcs to $500$ vertices and $5000$ arcs. We generated $30$ networks for each size / generator combination and performed an evolutionary search runs on each one of them. We found a correct results rate of $97.3\%$ for preferential attachment and $94\%$ for random. In the preferential attachment case, the precise solution with no bloat ($w=w_{\text{PA}}$) was found $92.7\%$ of the time. In the random network case, the precise solution with no bloat ($w=w_\text{ER}$) was found $76.7\%$ of the time. Interestingly, this result on a series of stochastic realisations of the ER and PA models is a strong indication that a real network which does not lead to the discovery of $w_{\text{ER}}$ or $w_{\text{PA}}$ obeys a more sophisticated morphogenesis process (see Supp. Info. for detailed results).

We then proceeded to experiment with seven datasets from a diverse
selection of real-world contexts: the neural network of a \hbox{C.
  Elegans} roundworm \cite{bib16, bib17}, a network of political blogs
\cite{bib18}, a software collaboration
network~(http://cpan-explorer.org/category/authors/, date of access:
10/03/2014), the power grid of the Western States of the USA
\cite{bib17}, a social network extracted from the neighbourhood of a
single Facebook user \cite{bib19}, a network of protein interactions
in Homo Sapiens \cite{bib20} and a word adjacencies network \cite{bib27}. The first three are directed while the latter four are undirected. 

\begin{figure}
\begin{center}
\includegraphics[scale=0.5]{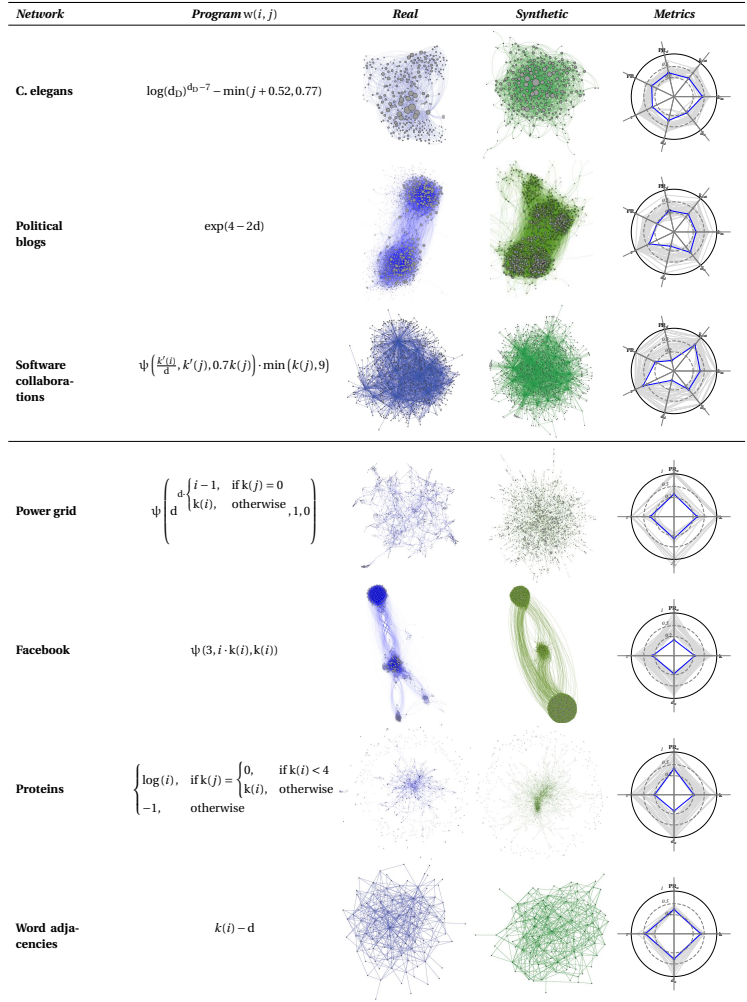} 
\end{center}
\caption{{\bf Overview of results for five datasets (one per row).} The columns, in order, represent: the generator expression, a visualisation of the synthetic and real networks and a radar graph showing each of the metrics -- the outer circle indicates the value of $1$, lower is better, best generator shown in blue, all others in grey. The symbols on the radars represent the various distribution distance measures employed in the fitness function: $k$, $k_{in}$ and $k_{out}$ for degree, in-degree and out-degree; $PR_d$ and $PR_u$ for directed and undirected PageRank; $d_d$ and $d_u$ for directed and undirected distance and $\tau$ for the triadic profile.}
\label{fig2}
\end{figure}

Figure~\ref{fig2} shows an overview of the results we obtained,
featuring the expression of the best program found after the {30}
evolutionary runs, as well as a comparison between the corresponding
synthetic network and original (target) network.

\begin{figure}
\begin{center}
\includegraphics[scale=0.53]{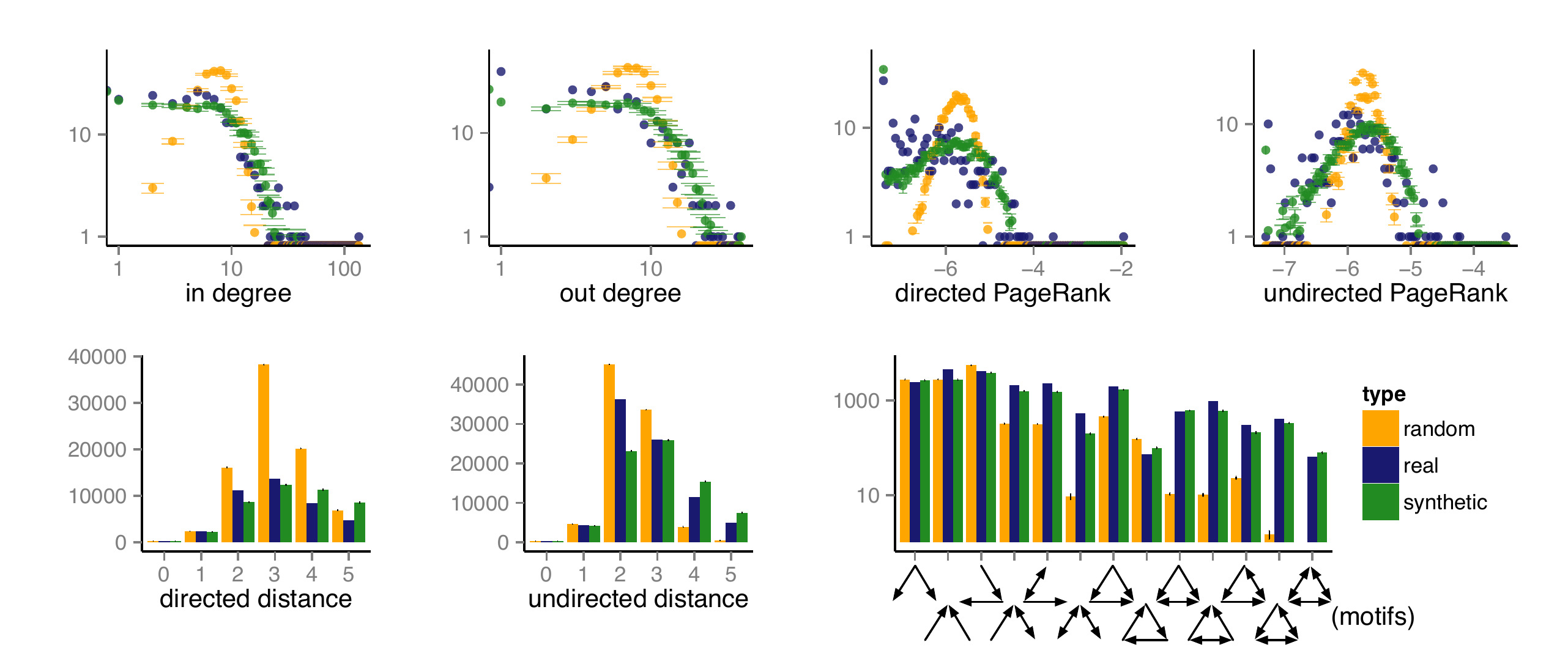} 
\end{center}
\caption{{\bf Real, synthetic and random distributions for the C.~Elegans neural network.} All $y$ axis represent frequencies. The synthetic and random series include error bars for a sample of $30$ runs.}
\label{fig3}
\end{figure}

Figure~\ref{fig3} focuses on \hbox{C. Elegans} and shows a comparison of the various distributions we use in our fitness function for the real network, a sample of {30} random networks with the same number of nodes and arcs, and a sample of {30} synthetic networks produced by the best generator we found for that network. Given the stochastic nature of the generative process, multiple runs of the same generator can produce different results. The figure shows that, in practice, variance is very small. Similar approximations were obtained for the other networks.

We provide an interpretation of each one of these generators in the \hyperlink{sup_mat}{supplemental materials}.

While these are high quality solutions according to the set of metrics we defined, another question is whether high-quality solutions generated by our method are similar to each other or represent completely different models. To investigate this issue we defined a process to quantify the similarity between two generators -- let us call them generators $w$ and $w'$. We produce a network using generator $w$ and, at each arc creation step, for each sample of candidate arcs, we also compute the probability of each candidate using generator $w'$. We then compute the mean distance between the probabilities assigned by generators $w$ and $w'$ to all the candidate arcs during the entire generative process. We thus get a {dissimilarity measure between generators} which we denote $d_{ww'}$. Conversely, we produce a network with generator $w'$ and compare the probabilities with the ones assigned by generator $w$, obtaining $d_{w'w}$. Finally, we consider the (generator) dissimilarity between $w$ and $w'$ to be $d = (d_{ww'} + d_{w'w}) / 2$.

\begin{figure}
\begin{center}
\includegraphics[scale=0.6]{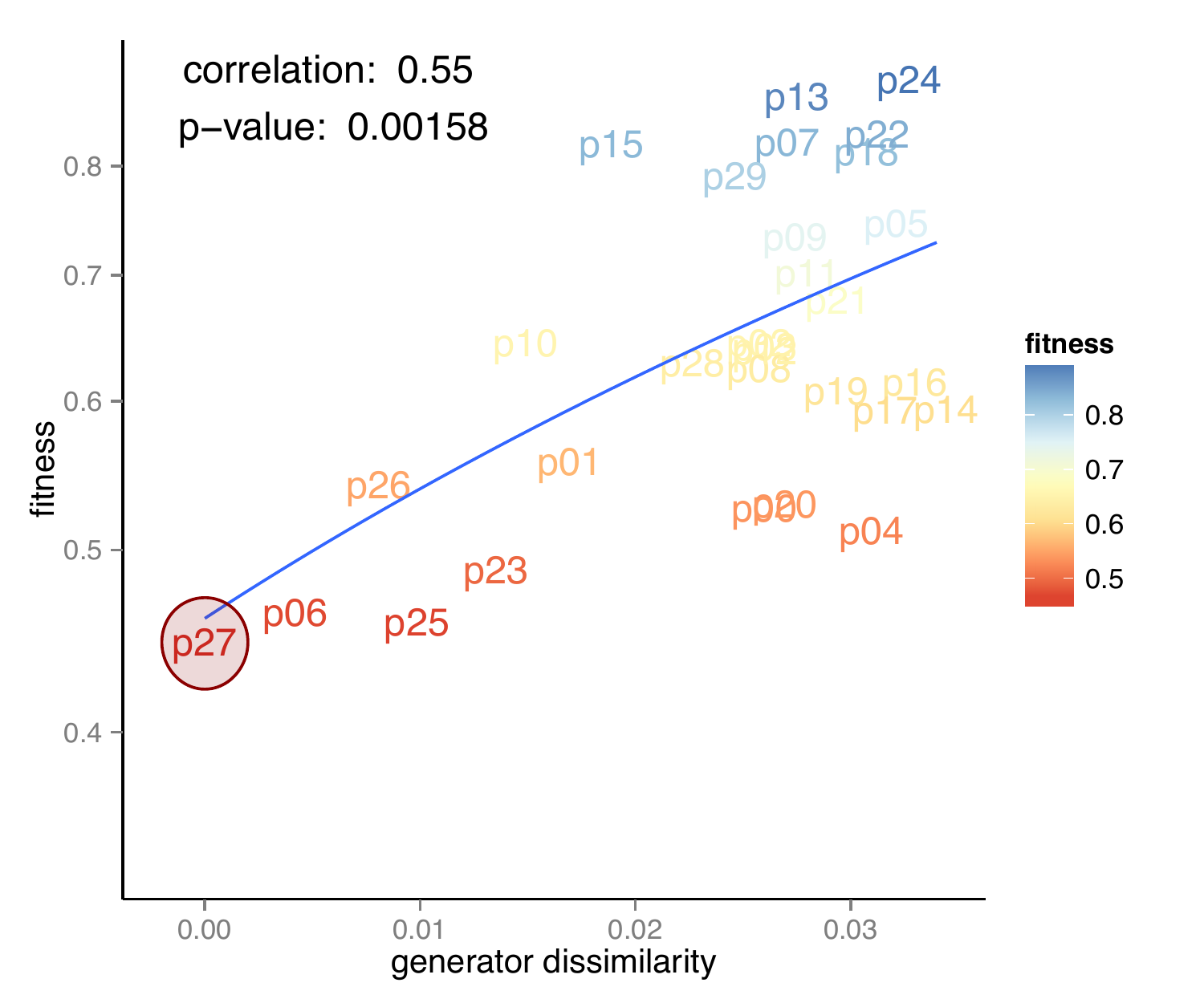} 
\end{center}
\caption{{\bf Similarity of generators.} Comparison between generator similarity to the optimal generator (p27) and fitness.}
\label{fig4}
\end{figure}

In the left panel of figure~\ref{fig4} we compare the (generator) dissimilarity between the optimal generator we found (p27) and all other generators obtained for \hbox{C. Elegans} with the fitness of these generators, \hbox{i.e.} max (network) dissimilarity on all metrics. The Pearson correlation indicates a strong relationship between fitness and similarity to the optimal generator. Furthermore, there is a significant probability that such a correlation exists ($p < 0.005$). 
On the right panel we also compare the distance with the mean dissimilarity in order to observe generators over all metrics, obtaining the same conclusions. 
The results we obtain provide compelling evidence that the closer the generators are to the best program in terms of fitness (at the network level), the closer they are in terms of the qualitative behaviour defined by their programs (at the link level), implying that this correlation further strengthens the plausibility of this generator.

Another point to note is that as program distance to the best solution increases, there is an increase in fitness variance. This is not surprising given that an increase in program distance corresponds to a decrease on the constraints on the space of possible programs. All the runs are subject to the same evolutionary pressure to decrease fitness, so it is likely that some become stuck in \emph{local minima} -- a common phenomena in heuristic search strategies. In fact, it is not possible to ever be sure that some result is not a \emph{local minima}, but this is also a limitation of the scientific method in general. However, we show that independent runs of our algorithm form a cluster of high quality results with respect to generator similarity. There is a degree of convergence on a consensus that facilitates the task of choosing between competing theories.

\section{Discussion}

We proposed a methodology to describe network generators and automatically manipulate them in order to assist in the discovery of plausible morphogenetic processes. We presented a number of reasons to be optimistic about this approach. The generator semantics proved to be expressive enough to represent growth processes that lead to structurally diverse networks and the evolutionary algorithm was able to find plausible generators for these different cases. The plausibility of the solutions is based on a comprehensive set of conventional metrics that reflect different aspects of a network's structure. The generators found are sufficiently succinct to have high explanatory power. Multiple runs of evolutionary search on the same network were shown to converge on similar solutions. Similarly, runs on stochastic realisations of canonical ER- and PA-based generators essentially led to the discovery of the correct original laws. More broadly, we believe our approach has a wide range of application domains where it could fruitfully guide scientists towards credible processes underlying the formation of the empirical networks they are trying to model.

There are many possible avenues to improve upon the method we propose. The vast array of techniques from the evolutionary computation and genetic programming bibliography could be employed. Larger populations and recombination operators may lead to higher quality results at the expense of computational tractability. Pareto optimisation may be used to explicitly select trade-offs between precision and solution complexity. Variables and operators for specific domains (e.g. spacial restrictions) can be introduced. In this work we strived for simplicity and generality, and to provide the scientific community with a tool that can be immediately useful but also serve as a baseline for further refinements.

\hypertarget{methods}{Methods}
\section{Methods}

\newcommand{\GC}{w_\text{c}}
\newcommand{\GO}{w_\text{o}}
\newcommand{\GS}{w_\text{s}}
\newcommand{\GR}{w_\text{r}}
The evolutionary algorithm maintains one or two generators at each
time: $\GO$ is the generator that produced the networks with the
lowest dissimilarity to the target network so far. $\GS$ is the
generator with the shortest program that produced a network with a
dissimilarity not more than $10\%$ worse than $\GO$. We refer to this
dissimilarity ratio as \emph{anti-bloat tolerance}. At any moment, it
is possible that $\GO = \GS$. This {procedure is meant to fight} bloat
--- the accumulation of needless complexity in generator programs \cite{bib12}. {The algorithm is initialised with a randomly created generator $\GR$ (see \hbox{Supp.} \hbox{Info.} for details). In the initial state $\GR = \GO = \GS$. For every evolutionary search generation, a parent generator is randomly selected from $\{\GO, \GS\}$. This parent generator is then cloned and mutated to produce the child generator $\GC$. Mutation consists of randomly selecting a sub-tree, removing it and replacing it with another randomly {selected sub-tree extracted from another randomly created tree}. $\GC$ is used to produce a synthetic network and the dissimilarity of this network to the target is computed. The dissimilarity and program length of $\GC$ is compared against $\GO$ and $\GS$, and $\GC$ will replace one or both if appropriate. The search will terminate once $\GO$ and $\GS$ remain unchanged for {a certain number of generations, we choose this number to be $1000$}. $\GS$ will be taken as the final result.}

Given the significant computational effort needed to generate a
network, we propose a strategy that limits the amount of such
generation steps. While it is common in evolutionary algorithms to use
large populations to prevent \emph{local minima}, this is not the only
possible strategy \cite{bib25}, nor is it guaranteed to work \cite{bib26}.

There are two parameters that introduce trade-offs in the search process: sample ratio and anti-bloat tolerance.

Sample ratio is a trade-off between generator accuracy (lower samples leading to more randomness against the linking preference defined by the generator) and computational effort (higher samples require more generator evaluations per link generation step).

Being $V$ the set of vertices, $s_r$ a predefined sampling ratio, $A$ the set of all possible arcs ($|A| = |V|^2$) and $A'$ the set of all arcs that do not currently exist in the network ($A' = \{a \in A | w_a = 0\}$, $w_a$ being the weight of arc $a$), we define a sample $S$ with $|S| = n = s_r \cdot |A|$ such that $S = \{s_1, ..., s_n\}$ with $s_i \in A'$.

In the experiments presented in this article, we do not allow duplicate or self-links. These restrictions could trivially be lifted if appropriate.

The value we propose was set sufficiently high to work with the smaller networks in our data set -- at some point, the sample becomes too small and the generators operate too randomly to lead to evolutionary improvement. Conversely, the sample size could be made smaller to reduce the computational effort for very large networks.

Anti-bloat tolerance is a trade-off between result quality and conciseness. Here we adjusted once and for all the value against our initial experiment, C. Elegans, and found $15\%$ to stall evolution and $5\%$ to lead to hard to interpret, bloated solutions. Without any further parameters adjustment, we then tested the algorithm against real and synthetic datasets, having found that this leads to perfect solution on the synthetic cases and robust results on the other $6$ real-world networks. It is possible that these parameters can be further optimised for specific cases or if more computational effort can be tolerated. However, in this work we strived to demonstrate the general applicability of the method.

The stop condition ($1000$ stable generations) and random tree generation parameters (detailed in \hbox{Supp.} \hbox{Info.}) are conventional genetic programming parameters and were set within ranges that are very common in the literature. Given the heuristic nature of genetic programming, it is impossible to avoid such parameters. Quoting ``A Field Guide to Genetic Programming'' \cite{bib14}:
\begin{quote}``It is impossible to make general recommendations for setting optimal parameter values, as these depend too much on the details of the application. However, genetic programming is in practice robust, and it is likely that many different parameter values will work.''\end{quote}

The quality and meaning of the results presented are not contingent on these parameters, as these only affect the search process itself. Further efforts on parametrisation may lead to higher quality results being found. We avoided such efforts to prevent a bias for our dataset. We propose that this increases credence on the general applicability of the method.

Ultimately, while we believe to have demonstrated the effectiveness of a heuristic search algorithm, this, of course, does not preclude refinements by further research.

\bigskip
\subsection*{Acknowledgments}
This work has been partially supported by the French National Agency of Research (ANR) through grants ``SIMPA'' ANR-09-SYSC-013-02 and ``Algopol'' ANR-12-CORD-0018. The authors are grateful to Chih-Chun Chen for proofreading the manuscript and to Jean-Philippe Cointet, John K. Clark, Jorge Tavares, Kim Jones, Marc Barthelemy, Russell K. Standish and Taras Kowaliw for reviewing the work and providing feedback.

\subsection*{Author contributions}
T.M. and C.R. contributed equally to this work.

\hypertarget{sup_mat}{Supplemental materials}
\subsection*{Supplemental Materials}
\url{http://www.nature.com/srep/2014/140905/srep06284/extref/srep06284-s1.pdf}

\end{document}